\documentclass[lettersize,journal]{IEEEtran}
\usepackage{amsmath,amsfonts}
\usepackage{algorithmic}
\usepackage{algorithm}
\usepackage{array}
\usepackage[caption=false,font=normalsize,labelfont=sf,textfont=sf]{subfig}
\usepackage{textcomp}
\usepackage{stfloats}
\usepackage{url}
\usepackage{verbatim}
\usepackage{graphicx}
\usepackage{booktabs}
\usepackage{graphicx}
\usepackage{cite}
\usepackage{booktabs}
\usepackage{multirow}
\usepackage{graphicx}
\begin{document}

\title{Self-Supervised Guided Segmentation Framework for Unsupervised Anomaly Detection}

\author{Peng~Xing,
	 Yanpeng~Sun,
	  Zechao~Li
        
	\IEEEcompsocitemizethanks{
		\IEEEcompsocthanksitem P. Xing,	Y. Sun, Z. li are with the School of Computer Science and Engineering, Nanjing University of Science and Technology, Nanjing 21094, China. E-mail: xingp\_ng@njust.edu.cn, yanpeng\_sun@njust.edu.cn,  zechao.li@njust.edu.cn. (Corresponding Author: Zechao Li)}
}


\markboth{Submission for IEEE TRANSACTIONS ON IMAGE PROCESSING}%
{Shell \MakeLowercase{\textit{et al.}}: IEEE TRANSACTIONS ON IMAGE PROCESSING}


\maketitle

\begin{abstract}
Unsupervised anomaly detection is a challenging task in industrial applications since it is impracticable to collect sufficient anomalous samples. In this paper, a novel Self-Supervised Guided Segmentation Framework (SGSF) is proposed by jointly exploring effective generation method of forged anomalous samples and the normal sample features as the guidance information of segmentation for anomaly detection. Specifically, to ensure that the generated forged anomaly samples are conducive to model training, the Saliency Augmentation Module (SAM) is proposed. SAM introduces a saliency map to generate saliency Perlin noise map, and develops an adaptive segmentation strategy to generate irregular masks in the saliency region. Then, the masks are utilized to generate forged anomalous samples as negative samples for training. Unfortunately, the distribution gap between forged and real anomaly samples makes it difficult for models trained based on forged samples to effectively locate real anomalies. Towards this end, the Self-supervised Guidance Network (SGN) is proposed. It leverages the self-supervised module to extract features that are noise-free and contain normal semantic information as the prior knowledge of the segmentation module. The segmentation module with the knowledge of normal patterns segments out the abnormal regions that are different from the guidance features. To evaluate the effectiveness of SGSF for anomaly detection, extensive experiments are conducted on three anomaly detection datasets. The experimental results show that SGSF achieves state-of-the-art anomaly detection results. 
\end{abstract}


\section{Introduction}

Unsupervised anomaly detection is a vital computer vision task, which includes anomaly classification and anomaly segmentation (localization) tasks. The goal of the anomaly classification task is to detect anomalous and normal samples, while the goal of the anomaly segmentation task is to locate the anomalous pixels. It has great potential in industrial scenes and medical diagnosis\cite {bergmann2019mvtec,bergmann2020uninformed,carrera2016defect,defard2021padim,zavrtanik2021draem,fei2020attribute,li2020superpixel,salehi2021multiresolution,WuL21,LeyvaSL17}. However, anomalous events are incidental and diverse resulting in the impracticality of collecting the sufficient anomalous samples \cite{gong2019memorizing,park2020learning}. Therefore, unsupervised anomaly detection is required to be trained by normal samples only, which is very challenging. 
\begin{figure}
	\centering
	\includegraphics[width=1\linewidth]{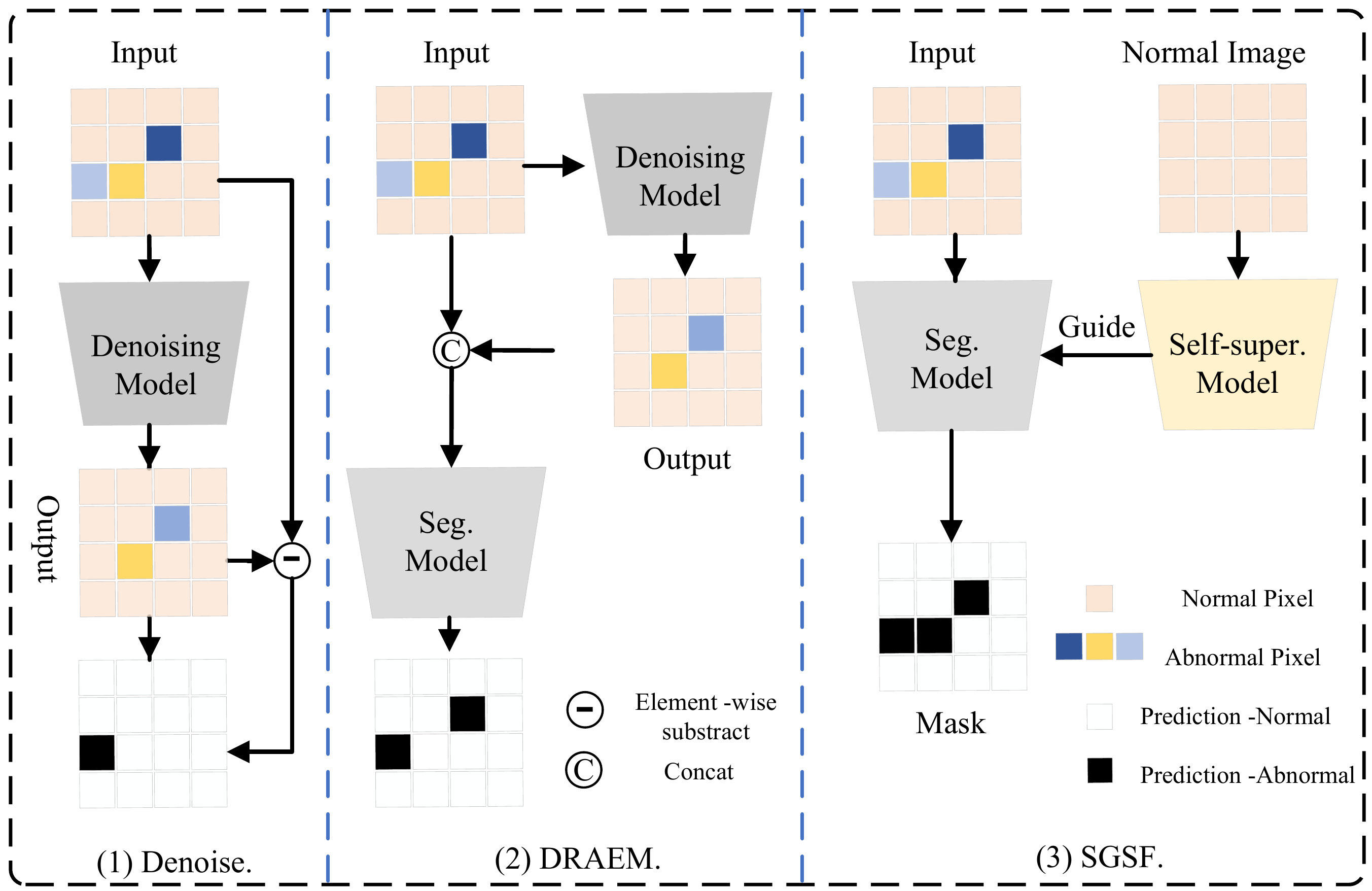}
	\caption{Comparison of different unsupervised anomaly detection methods. (1) Denoising-based method \cite{gong2019memorizing,bergmann2018improving}. (2) DRAEM method \cite{zavrtanik2021draem}. (3) The proposed framework (SGSF). The denoising network in methods (1) and (2) cannot accurately remove anomalies, resulting in inaccurate anomaly localization results. SGSF utilizes normal sample features to guide the segmentation network to locate anomalous regions.}
	\label{fig:kuangjia}
\end{figure}


Recently some methods have been studied to solve the anomaly detection task \cite{erfani2016high,li2003improving,ruff2018deep,defard2021padim}. The denoising models \cite{gong2019memorizing,hou2021divide},\cite{fei2020attribute} have been proposed to successfully reconstruct only normal samples rather than anomalous samples, as shown in Figure \ref{fig:kuangjia} (1). The reconstruction error between the original image and the reconstructed image is represented as the anomaly score of the sample, while the pixel-level error is represented as the anomaly segmentation image. Unfortunately, due to the lack of anomalous samples for training, the denoising network performs poorly. The denoising model degenerates into a reconstruction model in the face of real anomalous samples, and the anomalous regions are well reconstructed, which results in poor anomaly detection performance. 

\begin{figure*}
	\centering
	\includegraphics[width=1\linewidth]{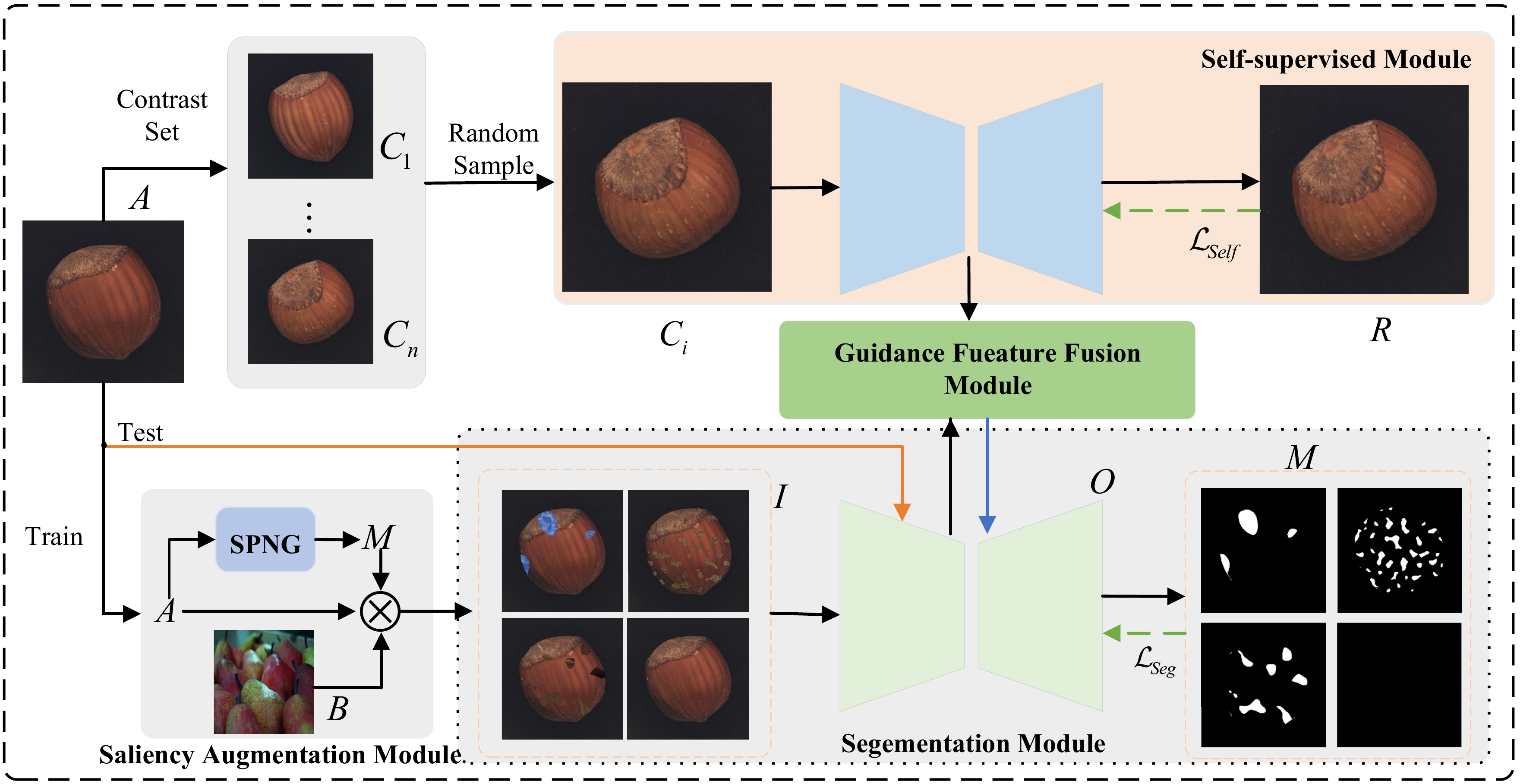}
	\caption{Illustration of the proposed SGSF. $A$ represents the input image. The contrast set $C$ is first generated from $A$,  then $C_i$ is randomly sampled from $C$ as the input of the self-supervised module. The self-supervised module extracts the noise-free and complete features as the guidance features for the segmentation module. During training, the image $A$ is input to the Saliency Augmentation Module (SAM) to generate the forged anomaly image $I$. The guidance fusion module fuses the guidance features with the features of $I$ into the decoder of the segmentation network. Finally, the segmentation map $O$ is obtained utilizing the decoder. The proposed SGSF uses self-supervised loss and segmentation loss to train the model. During testing, 
	the SAM is discarded and $A$ is fed directly into the segmentation module to achieve anomaly detection.}
	\label{fig:overallbig}
\end{figure*}
To solve the problem of imbalance between positive and negative samples due to the lack of anomalous samples for training, some methods \cite{zavrtanik2021draem,li2021cutpaste} are proposed to generate forged anomalous samples. CutPaste \cite{li2021cutpaste} generates forged anomalous samples by randomly cropping thin rectangular regions in normal samples, and then trains a classifier to detect the anomalous samples as well as obtains an anomaly localization map by the gradient of the classifier. However, the forged anomalous regions are only part of the normal sample and are rectangular in shape, which causes the forged anomalies to be too simple. Moreover, using the gradient of the loss function to find anomalous regions cannot be effective for tiny regions. Furthermore, DRAEM \cite{zavrtanik2021draem} is proposed to construct anomalous regions using global random Perlin noise \cite{Perlin85}. Besides, it uses the output of the denoising network to provide effective supervision information to locate anomalies by the segmentation network, as shown in Figure \ref{fig:kuangjia} (2). However, since the Perlin noise is global random noise, the generated anomalies will appear in the background areas of images. These forged anomalous samples may become noise samples to affect the model learning, which makes the model ineffective in real anomaly detection. Besides, the DRAEM method also depends on the performance of the denoising model. Since the denoising model trained by forged anomalous samples is difficult to guarantee the denoising quality, the output image of the denoising model is not guaranteed to provide effective guidance. Finally, anomalous samples are usually anomalous in tiny regions, and segmentation models have difficulty in solving the tiny object segmentation puzzle.

Towards this end, a novel Self-supervised Guided Segmentation Framework (SGSF) is proposed, which contains the Saliency Augmentation Module (SAM) and Self-supervised Guided Network (SGN), as shown in Figure \ref{fig:overallbig}. The goal of SAM is to forge anomalies in saliency regions of the image rather than background regions, which avoids to generate noisy samples affecting model learning. SAM introduces saliency map to generate saliency Perlin noise map. Then, an adaptive segmentation strategy is developed to segment the saliency Perlin noise map to generate a binarization mask. Finally, forged anomalous samples are generated by utilizing masks, auxiliary images and normal images.
The purpose of SGN is to fully exploit the normal samples features as guidance features to guide the segmentation module to locate anomalous regions, as shown in Figure \ref{fig:kuangjia} (3). To make the guidance information with noise-free and complete normal pattern information, the self-supervised module is proposed to extract effective features directly from normal samples. To accurately segment tiny regions, the guidance feature fusion module is proposed to guide the segmentation module with multi-scale features. The segmentation module with normal pattern features can detect anomalies that are different from the normal pattern.
  Extensive experiments are conducted on several anomaly detection datasets. The performance of SGSF over the state-of-the-art approches demonstrates its superiority.

 The main contributions of this paper can be summarized
 as follows:
 
\begin{itemize}
\item In this paper, we propose the novel Self-supervised Guided Segmentation Framework (SGSF) for unsupervised anomaly detection. The framework develops SAM to generate forged anomaly samples and SGN to directly extract normal samples as prior knowledge to guide segmentation module to locate anomalies.

\item The Saliency Augmentation Module(SAM) introduces a saliency map to generate saliency Perlin noise and utilizes an adaptive segmentation strategy to obtain the corresponding mask, which makes the generated forged anomaly samples more conducive to model learning.

\item The Self-supervised Guidance Network (SGN) is proposed, which directly extracts noise-free normal sample features as the prior knowledge, and utilizes the guidance feature fusion module to guide the segmentation module to detect anomalous regions that differ from the guidance features.

\end{itemize}
\par
The rest of this paper is organized as follows: Section II
is a brief review of related work. In section III, the proposed
method is introduced in detail. In section III, the experiment is introduced
in details. And section V is the conclusion.
\section{Related Work}
\subsection{Unsupervised Visual Anomaly Detection}
Anomaly detection is an important visual task. The diversity of anomalous samples leads to the impracticality of obtaining samples of all anomalies. Thus, anomaly detection models are only trained by normal samples \cite{zimek2012survey,golan2018deep,leung2005unsupervised}. It is very challenging for the model to detect anomalous scenes without learning real anomalous scenes. Early, anomaly detection aims to detect samples of unlearned categories, also calls OOD detection \cite{PangSCH21} or One-Class Classification \cite{ruff2018deep}. However, it is necessary to locate anomalous regions of abnormal images in industrial scenes and medical scenes \cite{FernandoGDSF22,zavrtanik2021draem,bergmann2019mvtec}. Existing methods include distribution-based methods, knowledge distillation-based methods, denoising network-based methods, and forged anomalous sample-based methods.

The distribution-based methods \cite{erfani2016high,li2003improving,ruff2018deep} first obtain the distribution of normal samples, then establish the relationship of the distribution. The ones that do not belong to such distributions are detected as anomalies. For example, DeepSVDD \cite{ruff2018deep} utilizes a deep learning model to map the original image space to a high-dimensional latent space. Since distribution of normal samples in the hidden space is constrained in fixed regions, the samples distributed in the remaining regions in the latent space are identified as anomalies. These methods perform well when the gap between normal and anomaly classes is large, but poor when the gap is small (e.g., leather scratch anomalies). The Padim  method \cite{defard2021padim} further attempts to address fine-grained anomaly detection. It models the distribution of each patch of the normal image by constructing multiple Gaussian mixture distribution models \cite{reynolds2009gaussian}. It allows regions that do not obey the distribution to be identified as outliers. However, It does not achieve good localization performance.

The knowledge distillation-based approach \cite{bergmann2020uninformed,salehi2021multiresolution} trains one or more student networks through a pre-trained teacher network and detects anomalies by the coding differences between the teacher and student networks for the samples. U-Std \cite{bergmann2020uninformed} utilizes multiple student networks and image patches as input to the network. However, it is inefficient in practice and can not achieve anomaly localization. MKDAD \cite{salehi2021multiresolution} uses only one student network and multi-scale feature distillation to encode normal samples closer together. It achieves anomaly localization, relying on the generated gradients to locate anomalous regions. However, performance is poor from the results.

The approaches based on the denoising network or autoencoder (AE) \cite{kingma2013auto} are proposed to successfully reconstruct normal samples and poorly anomalous samples \cite{fei2020attribute,bergmann2018improving,venkataramanan2020attention,tang2020integrating}. Abnormal regions are detected and localized by comparing the difference between the original image and the reconstructed image. These methods are implemented directly through AE \cite{bergmann2018improving}. However, due to the strong reconstruction ability of AE, they can reconstruct anomalous samples well even if there are no anomalous samples available in training. Some methods are proposed \cite{gong2019memorizing,park2020learning,wang2021cognitive,hou2021divide} to reduce the reconstruction ability of AE by introducing memory networks. MemAE \cite {gong2019memorizing} introduces a memory module after the encoder of the traditional AE. However, the simple memory module makes the performance poor in the face of fine-grained anomalies. The work \cite{park2020learning} further improves the original memory module and introduces a new loss function to improve the learning of the memory module. Unfortunately, these methods can still reconstruct unlearned objects well and the performance of anomaly detection remains poor.

The forged sample based approach is proposed to solve the problem of positive and negative sample imbalance in anomaly detection \cite{li2021cutpaste,zavrtanik2021draem}. After acquiring forged anomaly samples, fully supervised models can be used to solve the anomaly detection challenge.
 Cutpaste \cite{li2021cutpaste} generates forged abnormal samples that are added to the training set by using a part of the rectangular region of the original image to fill the forged region. Contrast learning is used to constrain the distance between positive samples to be as small as possible and the distance between negative samples to be as large as possible. However, the forged anomalies are not only rectangular in shape but also simple. Besides, using the gradient method to locate the anomalies, the pixel-level localization performance is poor. DRAEM\cite{zavrtanik2021draem} constructs more complex forged anomaly samples by generating complex masks and auxiliary images. It combines denoising and segmentation networks to achieve anomaly classification and anomaly localization. However, its generated anomalies are prone to background regions, which generate noisy samples and affect model learning. These methods cannot solve the challenge that the distribution gap between real and forged anomalies, which causes the model to be poor in real anomaly testing. Towards this end, this paper proposes a more reasonable method to generate forged anomalous samples as well as directly explore normal sample features to guide the segmentation network to detect anomalous regions.

\subsection{Segmentation methods In Anomaly detection}
Semantic segmentation models \cite{lin2017refinenet,long2015fully,RonnebergerFB15} show great potential for pixel-level localization problems. 
In medical anomaly detection task \cite{0004LYHCHS18,RonnebergerFB15,ChenYZAC16,reiss2021every,hsu2021darcnn}, excellent localization performance can be achieved by training end-to-end segmentation models using real anomaly samples and normal samples. For example, U-net \cite{RonnebergerFB15} was proposed to solve the problem of medical image segmentation. It uses a U-shape network structure to obtain contextual and positional information to segment the abnormal regions of lesions. Since the excellent localization performance of the segmentation network, some approaches \cite{xia2020synthesize,zavrtanik2021draem} introduce segmentation models to solve the anomaly localization challenge.
However, since only normal samples are available as training sets for anomaly detection, it is difficult to directly apply to the anomaly detection. Therefore, the premise of introducing the segmentation model is to construct forged anomalous samples in training. In \cite{xia2020synthesize}, the forged images are first generated by the GAN network as well as semantic segmentation network, and then the distillation network is used to compare the differences for OOD. However, it is only applicable to anomaly classification and difficult to achieve anomaly localization.
In this paper, we directly apply the segmentation network to implement anomaly localization and precisely locate anomalies based on the normal sample information as the prior knowledge.
\section{The Proposed Method }

In this section, we first present our overall architecture and
then describe each module in details.

\subsection{Overview}
\begin{figure}
	\centering
	\includegraphics[width=1\linewidth]{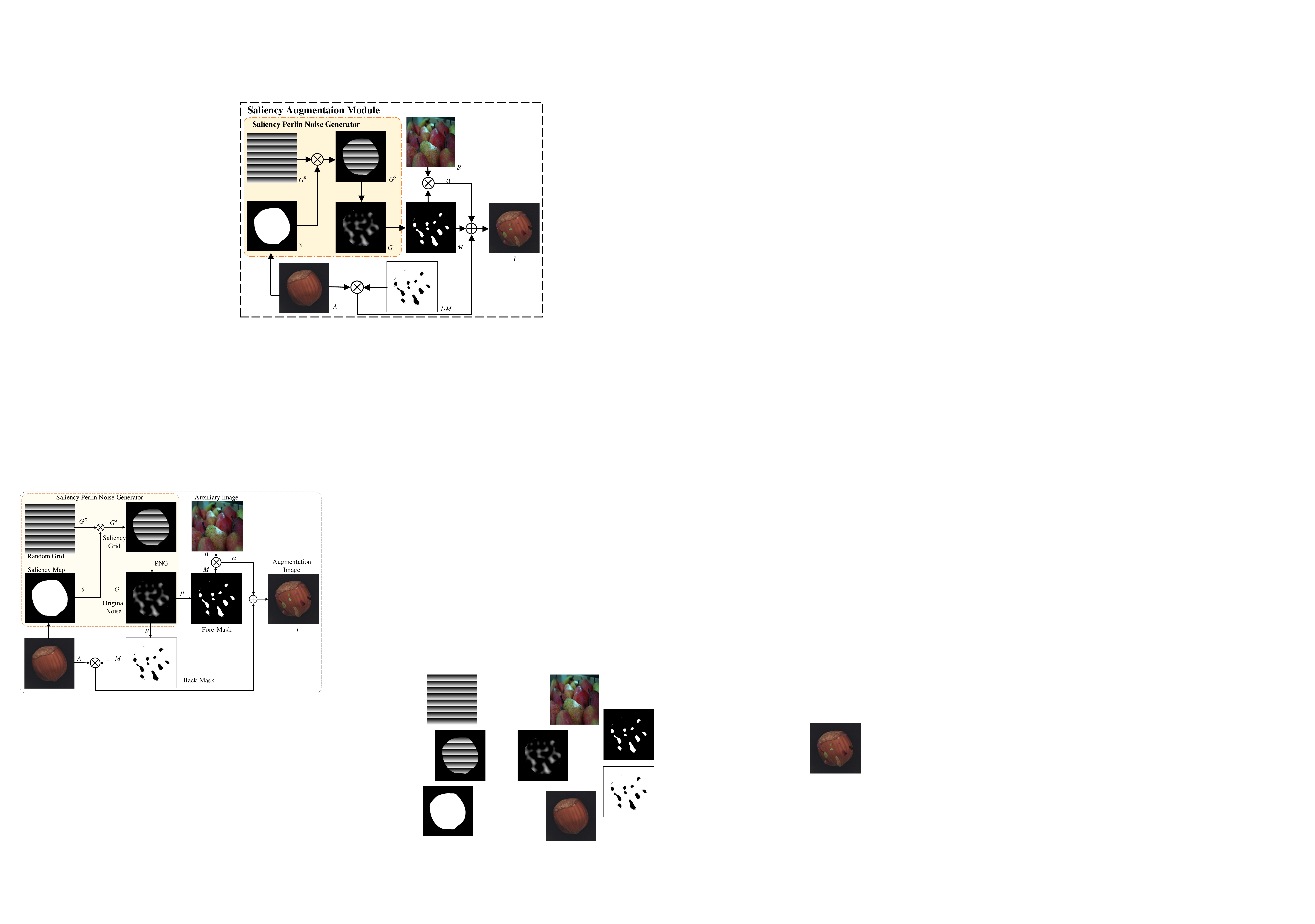}
	\caption{Illustration of the Saliency Augmentation Module. The input of the SAM consists of three parts: the original input image $A$, the auxiliary image $B$ and the saliency map $S$. The Saliency Perlin Generator generates the saliency Perlin noise $G$. The mask $M$ is obtained from the saliency Perlin noise $G$ using an adaptive threshold segmentation strategy. Finally, the original image $A$, auxiliary image $B$ and mask ($M$ and $1-M$) are processed to generate the forged anomaly image $I$. PNG represents the conventional Peilin noise generator, and $G^R$ represents the random grid.}
	\label{fig:aug}
\end{figure}

\par
The overall architecture of the proposed Self-supervised Guided Segmentation Framework (SGSF) is shown in Figure \ref{fig:overallbig}, which contains Saliency Augmentation Module (SAM) and Self-supervised Guided Network(SGN). SAM is developed to generate saliency Perlin noise by saliency maps. Subsequently, it can generate reasonable forged anomalous samples in the saliency region rather than the background region. In the training phase, the generated forged anomalous samples are served as training samples along with normal samples to train the segmentation model to locate the forged anomalous regions. In the test phase, the test samples are directly used as input to the segmentation module without input to SAM. SGN contains a self-supervised module, a guidance feature fusion module and a segmentation module. SGN is to provide noise-free guidance features to guide the segmentation module to locate anomalous regions. First, in the self-supervised module, generate the corresponding contrast set for each image, which is a set of normal images similar to the input image. Then, an image is randomly sampled from the contrast set and fed into the self-supervised module as input. The self-supervised module extracts clean and noise-free guidance features and feeds them to the guidance feature fusion module. The guidance feature fusion module fuses features extracted from the input image ($I$ or $A$) by the segmentation module with the guidance features at multi-scale. The result of the fusion serves as the input of the segmentation network. The segmentation network utilizes the guidance features as the prior knowledge and detects the features different from the guidance features as anomalous features.
The output of SGSF is the anomaly localization image. 
 Then, the anomaly score is obtained from the output anomaly localization image. In summary, SAM is proposed to generate forged anomalous samples, and SGN is proposed to utilize normal sample features to guide the segmentation module to locate anomalous regions.

\par

Formally, $A$ represents the input image and $y_A =$ $\left\{ {{{\rm{y}}^{ij}}|{y^{ij}} \in \{ 0\} } \right\}_{i = 1,j = 1}^{N,N}$ is pixel-level label, where $N$ represents the size of $A$, $y^{ij} =0 $ represents that the pixel is normal, $y^{ij} =1 $ represents that the pixel is abnormal. The obtained contrast set is $C = \{ C_1,C2,...,C_n\}$, $y_C =\left\{ {{{\rm{y}}^{ij}}|{y^{ij}} \in \{ 0\} } \right\}_{i = 1,j = 1}^{N,N} $, where $n$ denotes the number of samples in $C$. The sampled image is $C_i$ as the input of the self-supervised module. 
The auxiliary image is represented as $B$, $y_B =\left\{ {{{\rm{y}}^{ij}}|{y^{ij}} \in \{ 1\} } \right\}_{i = 1,j = 1}^{N,N} $. 
$R$ represents the output of the self-supervised module with the same size as $C_i$. The loss function of the self-supervised network is ${{\cal L}_{Self}}$. The image $A$ is forged by the SAM to obtain the set of samples $I$, and $y_I =\left\{ {{{\rm{y}}^{ij}}|{y^{ij}} \in \{0, 1\} } \right\}_{i = 1,j = 1}^{N,N} $. $O$ represents the output of the segmentation module for anomaly localization images. $M$ represents the forged mask. The loss function of the segmentation network is ${{\cal L}_{Seg}}$.

\subsection{Saliency Augmentation Module}
\par
SAM aims to generate reasonable forged anomalous samples that are close to the real anomaly distribution and suitable for model learning, which includes a Saliency Perlin Noise Generator (SPNG) and a Forged Sample Generator (FSG), as shown in Figure \ref{fig:aug}. 
Since DRAEM \cite{zavrtanik2021draem} relies on random Perlin noise to generate forged anomaly samples, the generated anomalies are prone to appear in the background and become noisy samples. To makes the generated anomaly mask in the saliency region rather than the background region, the proposed utilizes SPNG to generate saliency Perlin noise. The saliency Perlin noise is random noise generated only in the saliency region. FSG binarizes the saliency Perlin noise generated by SPNG with an adaptive threshold scheme to make the generated mask more diverse. Then, FSG utilizes the auxiliary samples, the mask generated by SPNG and the original image to generate the forged anomalous samples.

\begin{figure*}
	\centering
	\includegraphics[width=1\linewidth]{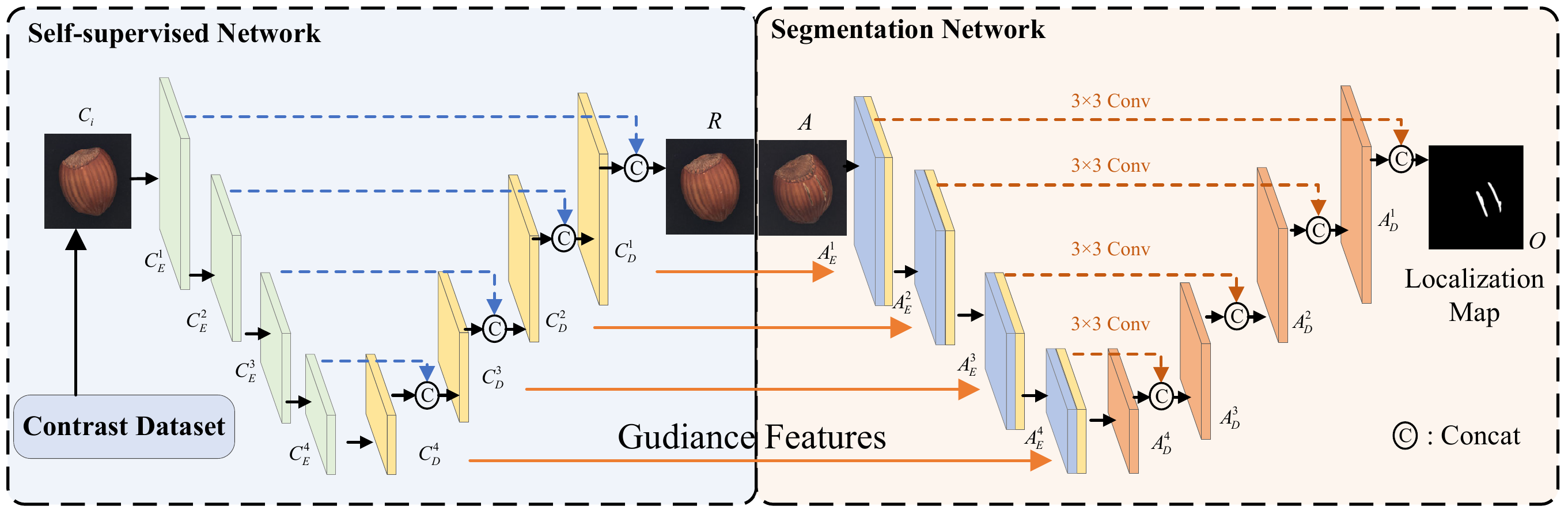}
	\caption{Illustration of the guidance feature fusion module and segmentation module. $C_i$ is randomly sampled from the contrast set C and used as the input of the self-supervised module. The features of $C_i$ are extracted by the self-supervised network as the guidance features ${C_D} = \{ C_D^i\} _{i = 1}^4$. The segmentation module extracts the features ${A_E} = \{ A_E^i\} _{i = 1}^4$ of the test image $A$. The guidance feature fusion module fuses $C_D$ with $A_E$ to achieve feature-level guidance. The fused results are subjected to 3 $\times$ 3 convolution to extract discriminative features, which utilize skip connections to guide the decoder to obtain anomalous localization map $O$.}
	\label{fig:network}
\end{figure*}

\par
Specifically, as shown in Figure \ref{fig:aug}, the input of SAM are the original image $A$, the saliency map $S$ extracted using the saliency method \cite{WuSH19}, and the auxiliary image $B$. 
First, similar to the conventional Perlin Noise Generator (PNG) \cite{zavrtanik2021draem}, a random grid $G^R$ is generated with the same size as the original image $A$. Second, the saliency map $S$ is Hadamard producted with with the original grid $G_R$ to obtain the saliency grid $G^S$, which is feed into the PNG to generate the original noise $G$. Third, by setting the threshold $\mu$, the foreground mask ($M$) and the background mask ($1-M$) are segmented. In this paper, an adaptive threshold scheme is developed associated with the saliency map is as follows:
\begin{equation}
\mu  = a + \frac{{sum(S)}}{{N*N}} \times b, \label{eq:1}
\end{equation}
where $[a, a + b]$ denotes the value domain of $\mu$, $sum(S)$ denotes the sum of the values of $S$. 
Finally, FSG fuses the image $A$, the auxiliary image $B$, the background mask $(1-M)$ and foreground mask ($M$) to generate a forged anomalous sample $I$. 
\begin{equation}
I = A \bullet (1 - M) + \alpha  \times (B \bullet M) + (1 - \alpha ) \times (A \bullet M),\label{eq:2}
\end{equation}
where $\alpha$ denotes the proximity of the simulation to the normal sample distribution, $\bullet$ is Hadamard product. The label of $I$ is composed of the labels of $A$, $B$, and $M$: $y_I =M \bullet y_B + (1-M) \bullet y_A$.

However, existing saliency detection methods identify the whole image as background when targeting textured images. Each pixel value on the saliency map is 0. We invert the saliency map to focus the saliency map to the full map. Formally,
\begin{equation}
	{G^S} = \left\{ {\begin{array}{*{20}{c}}
			{1,}&{sum(S) = 0,}\\
			{{G^S},}&{otherwise,}
	\end{array}} \right.
\end{equation}
where $sum(S)$ represents the sum of $S$ pixel values. 
Following \cite{zavrtanik2021draem}, the auxiliary image $B$ can be augmented by a series of augmentation methods to generate a wide variety of anomalous images.
\par

\begin{table*}[]
	\renewcommand\arraystretch{1.5}
	\caption{
		Comparison of the proposed method and unsupervised anomaly detection methods on MVTec dataset with AUROC (\%). The best results are marked in bold.}
	\resizebox{\textwidth}{!}{%
		\setlength{\tabcolsep}{1mm}{
			\begin{tabular}{@{}c|ccccccccccccccc|c@{}}
				\hline
				Methods & Carpet & Screw & Cable & Wood & Leather & Tile & Grid & Toothbrush & Zipper & Bottle & Hazelnut & Capsule & Metal nut & Pill & Transistor & MEAN \\ \hline
				GANomaly\cite{akcay2018ganomaly} & 82.1 & \textbf{100} & 71.1 & 92.0 & 80.8 & 72.0 & 74.3 & 70.0 & 74.4 & 79.4 & 87.4 & 72.1 & 69.4 & 67.1 & 80.8 & 78.2 \\
				MKDAD \cite{salehi2021multiresolution} & 79.3 & 83.3 & 89.2 & 94.3 & 95.1 & 91.6 & 78.0 & 92.2 & 93.2 & 99.4 & 98.4 & 80.5 & 73.6 & 82.7 & 85.6 & 87.7 \\
				U-Std \cite{bergmann2020uninformed} & 91.6 & 54.9 & 86.2 & 97.7 & 88.2 & 99.1 & 81.0 & 95.3 & 91.9 & 99.0 & 93.1 & 86.1 & 82.0 & 87.9 & 81.8 & 87.7 \\
				DAAD+ \cite{hou2021divide}& 86.6 & 98.7 & 84.4 & 98.2 & 86.2 & 88.2 & 95.7 & 99.2 & 85.9 & 97.6 & 92.1 & 76.7 & 75.8 & 90.0 & 87.6 & 89.5 \\
				RIAD \cite{zavrtanik2021reconstruction}& 84.2 & 84.5 & 81.9 & 93.0 & \textbf{100} & 98.7 & 99.6 & \textbf{100} & 98.1 & 99.9 & 83.3 & 88.4 & 88.5 & 83.8 & 90.9 & 91.7 \\
				MAD \cite{rippel2021modeling} & 95.5 & 81.2 & \textbf{94.0} & 97.6 & \textbf{100} & 97.4 & 92.9 & 95.8 & 97.9 & \textbf{100}  & 98.7 & 92.3 & 93.1 & 83.4 & 95.9 & 94.4 \\
				Cutpaste \cite{li2021cutpaste} & 93.1 & 86.3 & 80.6 & 98.6 & \textbf{100} & 93.4 & 99.9 & 98.3 & 99.4 & 98.3 & 97.3 & \textbf{96.2} & 99.3 & 92.4 & 95.5 & 95.2 \\
				Padim \cite{defard2021padim} & \textbf{99.8} & 85.8 & 92.7 & 99.2 & \textbf{100} & 98.1 & 96.7 & 96.1 & 90.3 & 99.9 & 92.0 & 91.3 & 98.7 & 93.3 & \textbf{97.4} & 95.5 \\ 
				\hline
				\textbf{SGSF} & 96.5 & 86.6 & 85.6 & \textbf{99.7} & \textbf{100} & \textbf{100} & \textbf{100} & \textbf{100} & \textbf{99.6} & 96.0 & \textbf{99.8} & 93.9 & \textbf{99.7} & \textbf{97.2} & 95.3 & \textbf{96.7} \\ \hline
			\end{tabular}%
	}}\label{TAB.1}
\end{table*}
\subsection{Self-supervised Guided Network}
\subsubsection{Self-supervised Module}
\par 
The purpose of the self-supervised module is to extract complete, noise-free, and normal semantic detailed features as guidance features, which serve as the prior knowledge to guide the segmentation module to identify anomalies. 

To ensure that the extracted features have complete and detailed feature information, the image reconstruction task is proposed as a self-supervised task in the self-supervised module. The image reconstruction task requires the model to first extract the image features and then reconstructs the original image by a decoder. Therefore, the output features by the encoder contain complete and detailed feature information. Besides, the input of the reconstruction model are normal samples to ensure that the extracted guidance features are normal semantic. In this paper, the guidance features are set as the output features of the encoder in the self-supervised network, which contains the complete feature information of the image while ensuring normal semantics.

Due to the large intra-class variation of the samples, even for normal samples, they are not similar in low-level features such as texture and color. If normal samples are randomly selected as the input for self-supervision model, it is difficult to extract features that can guide anomalous sample detection.
Therefore, this paper introduces the contrast set as the input of the self-supervised module. The contrast set refers to the set which samples from the normal samples that are similar to the input image $A$. The reason is to control the similarity of the guidance features and image $A$ in the normal region of semantic features with low-level features. Thus, the guidance features obtained by the segmentation module are more effective and guide the segmentation network to locate tiny anomalies by guided low-level features.
Specifically, the proposed approach train a network with AE, utilizing the output $q \in {\mathbb{R}^{512 \times 8 \times 8}}$ of last layer of the encoder as the features for measuring similarity. Then $q$ is flattened to be a vector $p_A \in \mathbb{R}^{512}$ by the mean-pooling layer. $p_A$ obtained from image $A$ is used to calculate the cosine similarity $T$ with the training images $p_{Train}$.

\begin{equation}
	T = \{ {T^1,T^2,...,T^j,...,T^m}{\rm{|}}{T^j} = \frac{{{p_A} \cdot p_{Train}^j}}{{\left\| {{p_A}} \right\| \times \left\| {p_{Train}^j} \right\|}}\} ,
\end{equation}
where $m$ is the number of training samples involved in constructing the contrast set. Then, $T$ is arranged as $T^\prime$ in descending order of similarity. Top $k$ values in $T^\prime$ corresponding to the original images are selected to compose the contrast set of $A$. 
\par
Importantly, the contrast set of image $A$ does not include $A$ itself, which can prevent the network to pay excessive attention to low-level features. In the training phase, randomly sample image from the contrast set are used as the input of self-supervised module. In the testing phase, the control set contains only the normal samples that are most similar to the input samples $A$.

\subsubsection{Guidance Feature Fusion Module and Segmentation Module}
\par 
The input to the guidance feature fusion module are the guidance features and the features extracted by the encoder of the segmentation module. It fuses them to achieve feature ``guidance" function, as shown in Figure \ref{fig:network}. To ensure the localization of tiny anomalous regions, a multi-scale guidance feature fusion method is utilized. 
\par 
Formally, the multi-scale features extracted by the encoder of the self-supervised network are ${C_E} = \{ C_E^i\} _{i = 1}^4$. The output of the decoder is ${C_D} = \{ C_D^i\} _{i = 1}^4$. The output of the encoder of the segmentation network is ${A_E} = \{ A_E^i\} _{i = 1}^4$ and the output of the decoder is ${A_D} = \{ A_D^i\} _{i = 1}^4$. $
C_E^i,C_D^i,A_D^i,A_E^i$ are same size, and $C_E^i \in {\mathbb{R}^{{2^{4 - i}}h \times {2^{4 - i}}w \times \frac{c}{2^{(4-i)}}}}.$
$C_D$ is defined as the guidance features. 

The guidance feature fusion module mainly implements the fusion of the guidance features with the features extracted by the segmentation module and the processing of the fusion features.
Taking $C_D^4$ as an example, 
the guidance features $C_D^4$ are first fused with $A_E^4$ to obtain a new tensor $F{ \in \mathbb{R}^{h \times w \times 2c}}$. In this paper, a simple, yet very effective fusion method is used, where two features are obtained by dimensional concatenation. $F$ is fed directly to the next layer of encoders, and the fused features obtained from the last layer of encoder are fed directly to the decoder.
It is well known that the introduction of skip connection can improve the performance of the segmentation model. Therefore, skip connection is introduced in the segmentation module of SGSF. To further compare the extracted features with the guidance features and to obtain more discriminative feature information for segmentation, the fused features are fed into the decoder after the convolutional layer and then utilizing the skip connection. The processed feature is $F'{ \in \mathbb{R}^{h \times w \times c}}$, which is combined with the decoder to obtain the anomaly localization image $M$.

The pixel values in the localization image $M$ are used as the probability of each pixel to be detected as anomalous. Following the method \cite{zavrtanik2021draem}, $M$ is processed through the mean-pooling layer to aggregate local anomaly information. Besides, the proposed uses the maximum value of $M$ rather than the average value as the anomaly score to remove noise that brings negative effects.

\subsection{ Loss Function}
The proposed contains self-supervised loss ${{\cal L}_{Self}}$ and segmentation loss ${{\cal L}_{Seg}}$.
In the reconstructed network of the self-supervised module, the $l_2$ loss is introduced. Besides, the $SSIM$ loss function\cite{wang2004image} is also introduced to better enhance the reconstruction so that the output guidance features are completely noise-free normal sample features. Follow \cite{BergmannLFSS19}, The $SSIM$ loss is obtained from Equation (\ref{equ6}). 
It is expressed as follows:
\begin{equation}
{\cal L}_{l_2} = {\left\| {{y_C} - {y_R}} \right\|_2},
\end{equation}

\begin{equation}
{{\cal L}_{SSIM}} = \frac{1}{{N \times N}}\sum\limits_{i = 1}^N {\sum\limits_{j = 1}^N {1 - SSIM_{(i,j)}^{k}{{({y_C},{y_R})}}}, } \label{equ6}
\end{equation}
where $ SSIM_{(i,j)}^{k}{{({y_C},{y_R})}}$ is denotes as the structural similarity index of a block of size $k\times k$ centered at $(i,j)$. The reconstruction loss is denoted as:
\begin{equation}
	{\cal L }_{Self}={\cal L}_{l_2}+{\cal L}_{SSIM}.
\end{equation}
For the segmentation module, the Focal loss \cite{LinGGHD17} is introduced to alleviate the imbalance between the normal pixels and anomalous pixels as well as improve the stability and effectiveness of the model. 
The Focal loss ${\cal L}_{Seg}$ is expressed as follows:
\begin{equation}
{\cal L}_{Seg} = \left\{ {\begin{array}{*{20}{c}}
		{ - {{(1 - p)}^\tau }\log (p),}&{y_M^{ij} = 1,}\\
		{ - {p^\tau }\log (1 - p),}&{y_M^{ij} = 0.}
\end{array}} \right.\label{equ8}
\end{equation}
where $p$ represents the probability of pixel $(i,j)$ prediction.
 Finally, the total loss of the framework is expressed as:
\begin{equation}
	{\cal L }=\lambda \times{\cal L }_{Self}+{\cal L }_{Seg} , \label{eq:11}
\end{equation}
where $ \lambda $ is set to measure the importance of the two loss functions.
\section{EXPERIMENTS}

\begin{table*}[]
	\centering
	\caption{Comparison of the proposed method and unsupervised anomaly detection methods on DAGM dataset with AUROC (\%). $N$ in MSDD indicates that MSDD uses $N$ pixel-level labeled anomaly samples and the all image level labeled anomaly samples. The best results are marked in bold.}
	\renewcommand\arraystretch{1.4}
	\setlength{\tabcolsep}{4mm}{%
		\begin{tabular}{@{}c|c|cccccc|cc@{}}
			\hline
			\multirow{2}{*}{Type}                  & \multirow{2}{*}{Methods}                           & \multicolumn{6}{c|}{10 categories}    & \multicolumn{2}{c}{6 categories}   \\ \cmidrule(l){3-10} 
			&              & AUROC  & AP   & F1   & TPR  & TNR  & CA   & CA   & mACC   \\ \hline
			\multirow{2}{*}{Supervised} &
			CCNN \cite{racki2018compact} &
			99.6 &
			- &
			- &
			\textbf{99.9}&
			99.4 &
			\textbf{99.6} &
			99.2 &
			99.4 
			\\
			&ENSDD\cite{lin2020efficient} &	- &	99.0 &- &99.4 &\textbf{99.9} &-&-&
			\textbf{99.8}  \\ \hline
			\multirow{4}{*}{Weakly-supervised} & CADN-W18  \cite{zhang2021cadn}                                         & -    & -    & 63.2 & - & - & 86.2 & -             & -               \\
			& CADN-W18(KD)\cite{zhang2021cadn}  & -    & -    & 65.8 & -    & -    & 87.6 & -    & -      \\
			& MSDD(N=0) \cite{BozicTS21} & 74.0 & 86.1 & 74.6 & - & - & 89.7 & 85.4          & 81.4            \\ 
			& MSDD(N=5) \cite{BozicTS21} & 91.5 & 94.9 & 92.3 &   &   & 92.9 & 88.1          & 91.6            \\ \hline
			
			\multirow{8}{*}{Unsupervised}      
			& F-AnoGAN   \cite{schlegl2019f}                                        & 57.5 & 19.5 & 27.8 & - & - & 79.7 & 81.7          & 54.6            \\
			& U-std \cite{bergmann2020uninformed}        & 86.4 & 66.8 & 67.1 & -    & -    & 84.3 & 79.3 & 78.5   \\
			& Staaer  \cite{staar2019anomaly}     & 83.0 & -    & -    & -    & -    & -    & -    & -      \\
			& RIAD \cite{zavrtanik2021reconstruction}       & 78.6 & -    & -    & 79.2 & 69.1 & 70.4 & -    & -      \\
			& MAD   \cite{rippel2021modeling}       & 82.4 & -    & -    & 78.7 & 85.7 & 66.2 & -    & -     \\
			& PaDim  \cite{defard2021padim}      & 95.0 & -    & -    & 83.3 & 97.5 & 95.7 & -    & -      \\
			& DRAEM  \cite{zavrtanik2021draem}      & 99.0 & -    & -    & 96.5 & 99.4 & 98.5 & -    & -     \\
			&  \textbf{SGSF} &
			\textbf{99.9} &
			\textbf{99.7} &
			\textbf{98.2} &
			98.1 &
			99.8 &
			99.5 &
			\textbf{99.6} &
			99.1   \\ \hline

		\end{tabular}%
	}\label{TAB.2}

\end{table*}
\begin{table}[]
	\centering
	\caption{Comparison of SGSF and unsupervised anomaly detection methods on KolektorSDD2 dataset with AUROC (\%). The best results are marked in bold.}
	\renewcommand\arraystretch{1.5}
	\setlength{\tabcolsep}{5mm}{
		\begin{tabular}{@{}c|ccl@{}}
			\hline
			Type       & Methods   &  AP$_{det}$ &  \\ \hline
			\multirow{2}{*}	{Weakly-supervised}& MSDD(N=0) \cite{BozicTS21}  & 73.3                    &  \\
			& MSDD(N=16) \cite{BozicTS21} & 83.2                    &  \\ \hline
			\multirow{3}{*}	{Unsupervised } & F-AnoGan \cite{schlegl2019f}            & 55.0                    &  \\
		    & U-Std  \cite{bergmann2020uninformed}            & 65.3                    &  \\
			
			& \textbf{SGSF}               & \textbf{86.3}           &  \\ \hline
		
		\end{tabular}%
	}\label{TAB.3}
\end{table}
\par
In this section, we introduce the experimental datasets, the experimental setup, as well as the performance of anomaly classification and anomaly localization. In addition, we conduct a number of ablation experiments to demonstrates  the impact of each module.
\subsection{Datasets}
\begin{table*}[]
	\renewcommand\arraystretch{1.5}
	\caption{Anomaly localization performance of the proposed method and other anomaly detection methods on the MVTec dataset.}
	\resizebox{\textwidth}{!}{%
		\setlength{\tabcolsep}{1mm}{
			\begin{tabular}{c|c|ccccccccccccccc|c}
				\hline
				Metric                 & MEthods  & Carpet        & Screw         & Cable         & Wood          & Leather       & Tile          & Grid          & Toothbrush    & Zipper        & Bottle        & Hazelnut      & Capsule       & Metal\_nut    & Pill          & Transistor    & MEAN          \\ \hline
				\multirow{7}{*}{AUROC$_{loc}$} & U-Std    & 93.5          & 97.4          & 91.9          & 92.1          & 97.8          & 92.5          & 89.9          & 97.9          & 95.6          & 97.8          & 98.2          & 96.8          & 97.2          & 96.5          & 73.7          & 93.9          \\
				& RIAD     & 96.3          & \textbf{98.8} & 84.2          & 85.8          & 99.4          & 89.1          & 98.8          & 98.9          & 97.8          & 98.4          & 96.1          & 92.8          & 92.5          & 95.7          & 87.7          & 94.2          \\
				& Cutpaste  & 98.3          & 96.7          & 90.0          & 95.5          & \textbf{99.5} & 90.5          & 97.5          & 98.1          & \textbf{99.3} & 97.6          & 97.3          & 97.4          & 93.1          & 95.7          & 93.0          & 96.0          \\
				& SPADE \cite{cohen2020sub}   & 97.5          & 98.9          & \textbf{97.2} & 88.5          & 97.6          & 87.4          & 93.7          & 97.9          & 96.5          & 98.4          & 99.1          & \textbf{99.0} & 98.1          & 96.5          & 94.1          & 96.5          \\
				& PaDim    & \textbf{99.0} & 94.4          & 96.7          & 94.1          & 99.0          & 94.1          & 98.6          & 98.8          & 98.4          & 98.2          & 98.1          & 98.6          & 97.3          & 95.7          & \textbf{97.6} & \textbf{97.4}        \\
				& DRAEM    & 95.5          & 97.6          & 94.7          & 96.4          & 98.6          & 99.2          &\textbf{99.7}          & 98.1          & 98.8          & \textbf{99.1} & 99.7          & 94.3          & \textbf{99.5} & 97.6          & 90.9          & 97.3 \\
				& \textbf{SGSF}    & 97.4          & 97.2          & 93.3          & \textbf{97.3} & 98.9          & \textbf{99.7} & 99.2 & \textbf{99.6} & 98.2          & 96.3          & \textbf{99.5} & 97.4          & \textbf{99.5} & \textbf{99.5} & 84.7          & 97.2\\ \hline
				\multirow{6}{*}{AP$_{loc}$}    
				& U-Std    & 52.2          & 7.8           & 48.2          & 53.3          & 40.9          & 65.3          &  10.1         & 37.7          & 36.1          & 74.2          & 57.8          & 25.9          & 83.5          & 62.0          & 27.1          & 45.5          \\
				& RIAD     & 61.4          & 43.9          & 24.4          & 38.2          & 49.1          & 52.6          & 36.4         & 50.6          & 63.4          & 76.4          & 33.8          & 38.2          & 64.3          & 51.6          & 39.2          & 48.2          \\
				& PaDim    & 60.7          & 21.7          & 45.4          & 46.3          & 53.5          & 52.4          & 35.7          & 54.7          & 58.2          & 77.3          & 61.1          & 46.7          & 77.4          & 61.2          & \textbf{72.0} & 55.0          \\
				& DRAEM    & 53.5          & \textbf{58.2} & 52.4          & 77.7          & \textbf{75.3} & 92.3          & 65.7         & 44.7          & 81.5          & \textbf{86.5} & \textbf{92.9} & 49.4          & 96.3          & 48.5          & 50.7          & 68.4          \\
				& \textbf{SGSF}     & \textbf{70.5} & 43.8          & \textbf{62.0} & \textbf{84.4} & 73.2          & \textbf{97.2} & \textbf{66.9} & \textbf{78.9} & \textbf{83.2} & 81.5          & 83.7          & \textbf{51.6} & \textbf{96.8} & \textbf{93.2} & 40.1          & \textbf{73.8} \\ \hline
			\end{tabular}%
	}}\label{tab4}
\end{table*}

\emph{\textbf{(1) MVTec \cite{bergmann2019mvtec} :}}
MVTec is a challenging anomaly detection dataset. It contains a series of industrial real abnormal and normal images. The MVTec dataset includes 15 different categories, containing both texture images (e.g.,``leather") and non-textured images (e.g., ``cable"). Anomaly classification and anomaly localization can be evaluated. In the training phase, only normal samples can are for training, and different classes of anomalies need to be detected. All images are high-resolution images and some anomalous regions are tiny, such as ``broken screw".
\par
\emph{\textbf{(2) KolektorSDD2 \cite{BozicTS21} :}}
KolektorSDD2 is a surface anomaly detection dataset from real scenes. The dataset consists of color images of defective production items, captured by a vision inspection system. The images in the dataset are captured in a controlled environment and are similar in size, about 230 $\times$ 630. The dataset is complex and accurately annotated. In the original dataset, the training set contains a large number of abnormal samples. In this work, we first process the original dataset so that it contains only normal samples for training. Finally, the number of training samples are 2085 and the test samples are 1004. The anomalies include scratches or large surface defects with different shapes, sizes and colors.
\par
\emph{\textbf{(3) DAGM \cite{AlbelwiM17} :}} 
 DAGM is a well known dataset for surface defect detection. It contains ten different categories of computer-generated gray images of surfaces and various defects. Early, only six categories of data were publicly evaluated, and subsequently four categories were publicly available. The anomaly label of this dataset is inaccurate. Therefore, only the anomaly classification performance of the DAGM dataset is evaluated. 
 \subsection{Baselines and Evaluation Metric}
 \emph{\textbf{(1)Baselines :}}
 SGSF is compared with state-of-the-art
 deep methods including fully supervised methods, i.e., CCNN \cite{racki2018compact} and ENSDD\cite{lin2020efficient}, weakly supervised methods, i.e., CADN-W18 \cite{zhang2021cadn}, CADN-W18 (KD) \cite{zhang2021cadn}, and MSDD \cite{BozicTS21}, as well as unsupervised methods, i.e., GANomaly \cite{akcay2018ganomaly}, MKDAD \cite{salehi2021multiresolution}, U-Std \cite{bergmann2020uninformed}, DAAD+ \cite{hou2021divide}, RIVAD \cite{zavrtanik2021reconstruction}, MAD \cite{rippel2021modeling}, Cutpaste \cite{li2021cutpaste}, Padim \cite{defard2021padim}, and SPADE \cite{cohen2020sub}. Fully supervised methods use all the normal samples with pixel-level labels. The weakly supervised method uses image-level labeled abnormal samples and all normal samples, or uses a part of pixel-level labeled abnormal samples and all normal samples.  The unsupervised method only uses normal samples. The MKDAD, U-Std and Cutpaste use the pretrained model. The Cutpaste and DRAEM use forged anomalous samples.

 \begin{table}[]
 	\centering
 	\renewcommand\arraystretch{1.5}
 	\caption{Ablation studies on self-supervised module and SAM.}
 	
 	\setlength{\tabcolsep}{1.6mm}{
 		\begin{tabular}{@{}c|c|cccc@{}}
 			\hline
 			& \textbf{}     & (1)  & (2)  & (3)  & SGSF  \\ \hline
 			\multirow{2}{*}{\begin{tabular}[c]{@{}c@{}}Self-supervised\\ Module\end{tabular}} & Reconstruction task &      & \checkmark &      & \checkmark    \\
 			& Denoising task &      &      & \checkmark    &      \\ \hline
 			\multirow{2}{*}{\begin{tabular}[c]{@{}c@{}}SAM \end{tabular}}    & SPNG              & \checkmark    &           & \checkmark    & \checkmark    \\
 			& PNG           &      & \checkmark    &      &      \\ \hline
 			\multirow{3}{*}{Metric}                                                           & AUROC$_{det}$.       & 94.7 & 95.1      & 94.8 & 96.7 \\
 			& AUROC$_{loc}$.   & 94.3 & 95   & 94.1 & 97.2 \\
 			& AP$_{loc}$.      & 56.7 & 67.7 & 66.3 & 73.8 \\ \hline
 		\end{tabular}%
 	}\label{Tabel5}

 \end{table}
\par
 \emph{\textbf{(2) }}
Following previous work \cite{sabokrou2018deep,salehi2021multiresolution,bergmann2019mvtec}, image-level and pixel-level receivers operating characteristic curve and corresponding area under the curve (AUROC) were introduced to evaluate the performance of anomaly classification and anomaly localization. The metric of AUROC does not reflect the true performance in evaluating the localization performance. The reason is that the false positive rate is mainly determined by a-priori very high number of normal pixels \cite{zavrtanik2021draem}. However, the percentage of anomalous pixels is small in the test images, which makes the overall false positive rate performs at a low level. Therefore, the pixel-level average precision (AP) metric is introduced. It evaluates the accuracy of the overall pixel discrimination and evaluates the localization performance together with the pixel-level AUROC. $AUROC_{det}$, $AUROC_{loc}$ and $AP_{loc}$ represent AUROC for anomaly image classification, AUROC for anomaly image localization, and AP for anomaly image localization, respectively.
Following \cite{BozicTS21}, on the DAGM dataset, the image level AUROC, AP, F1, TPR, TNR, ACC, and mACC $= (TPR + TNR) / 2$ are introduced to evaluate the performance of anomaly classification.

\subsection{Implementation Details}
\par 
\emph{\textbf{(1) Sets:}} The number of contrast set $n$ is set to be 5. The auxiliary image $B$ in the Saliency Augmentation module is randomly sampled from the image set DTD\cite{cimpoi2014describing}. We follow the image augmentation method in the DRAEM and adjust the ratio of forged abnormal images to normal images to $3 : 1$ during the training phase.

\par 
\emph{\textbf{(2) Experimental Setting: }}
The proposed model is trained on single RTX TiTAN GPU using PyTorch. It is trained with a
mini-batch size of 16 using the Adam optimizer. The
learning rate is set to $1 \times 10^{-4}$. The learning rate decays to 0.1 times the current one at $800 \times $$[0.5, 0.7, 0.9]$ iterations, respectively. The hyperparameter $a$ is set to 0.4, $b$ is set to 0.2 in Eq. (\ref{eq:1}) and the input image size is set to 256 $\times$ 256. In Eq. (\ref{eq:2}) $a \in  [ 0.1, 1 ]$. In Eq. (\ref{eq:11}) $\lambda$ is set to 1 and in Eq. (\ref{equ8}) $\tau $ is set to 2.

\subsection{Comparison With State-of-the-Art Methods}

\begin{figure*}
	\centering
	\includegraphics[width=1\linewidth]{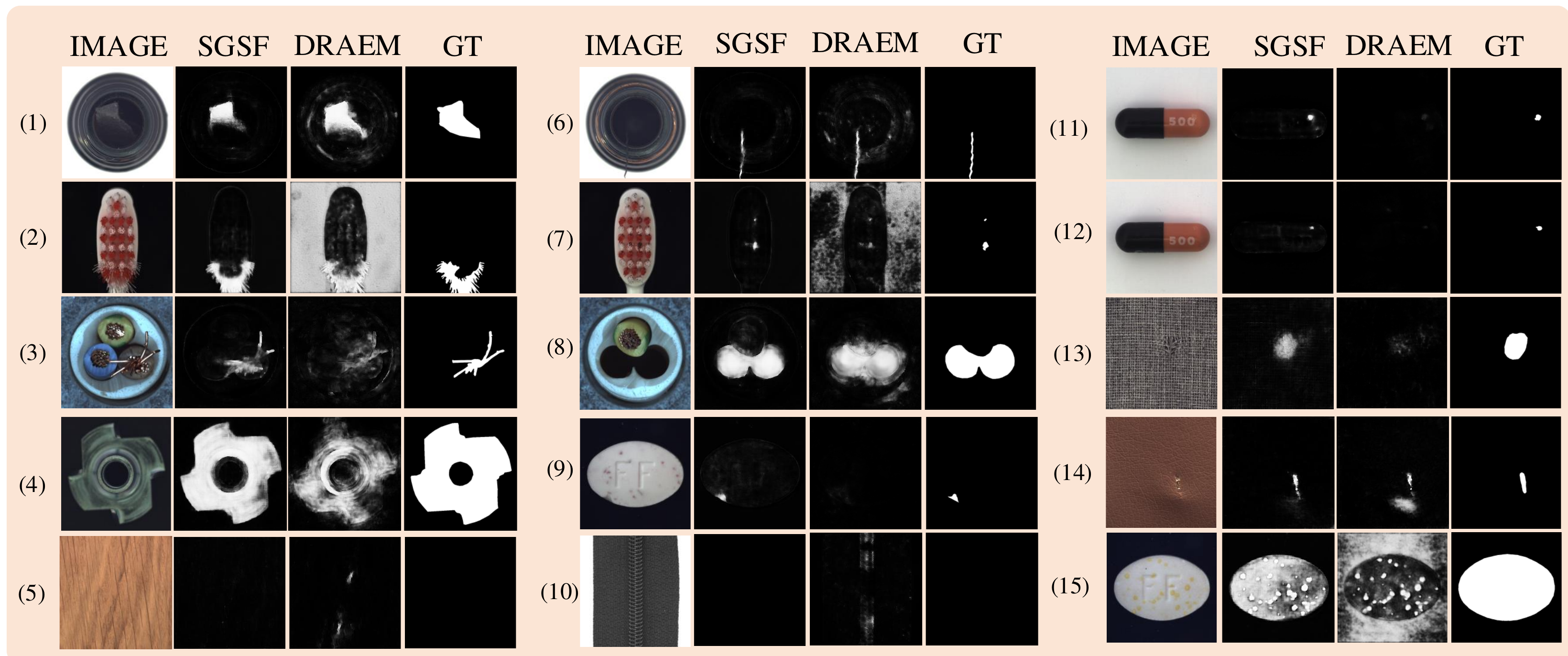}
	\caption{Localization results of the proposed SGSF with recent DRAEM method. ``IMAGE" denotes the test image. ``SGSF'' denotes the results of proposed method. ``DRAEM" denotes the results of DRAEM. ``GT" represents groundtruth.}
	\label{fig:compare}
\end{figure*}
\begin{figure}
	\centering
	\includegraphics[width=1\linewidth]{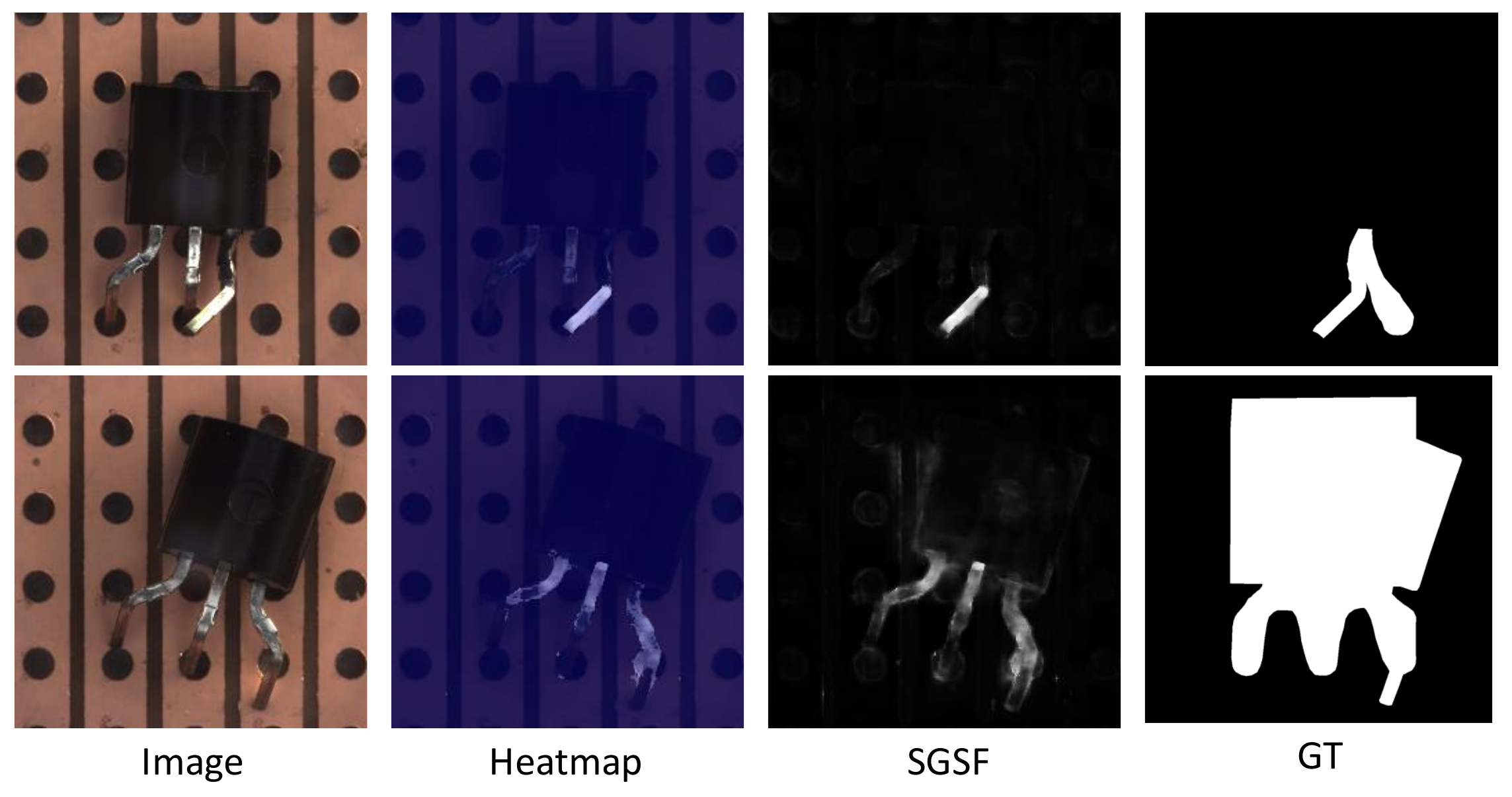}
	\caption{Illustration of the localization map and Ground Truth of some samples in the ``Transistor" category. These images need to label not only the regions where anomalies appear, but also the regions where components are missing.}
	\label{fig:tu7}
\end{figure}
\begin{figure}
	\centering
	\includegraphics[width=1\linewidth]{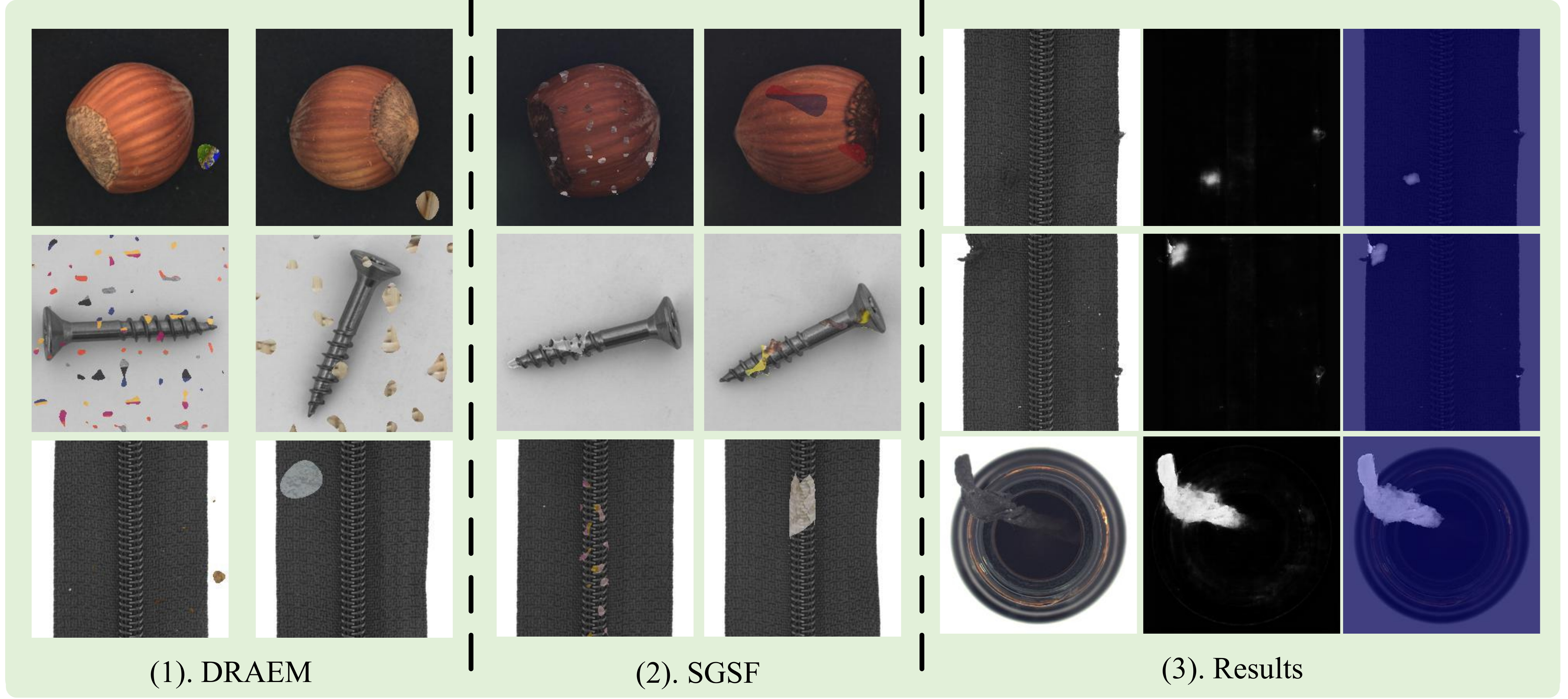}
	\caption{The forged anomalous samples generated by DRAEM method (1) and SAM (2) are illustrated. (3) shows that our method can locate the anomalous regions beyond the saliency region.}
	\label{fig:figure3}
\end{figure}

\emph{\textbf{(1) Anomaly image classification: }}
For the anomaly classification task, we evaluate the performance of the proposed method on MVTec, DAGM, and KolektorSDD2 datasets. 

In Table \ref{TAB.1}, we compare the anomaly classification performance of SGSF and unsupervised anomaly detection methods on MVtec dataset.
The experimental results show that SGSF is competitive, achieving mean $AUROC_{det}$ 96.7\%. Specifically, SGSF achieves $AUROC_{det} $100\% in four categories, such ``leather" and ``Toothbruth", outperforming the remaining unsupervised methods. Meanwhile, SGSF shows the best performance in 9 categories of MVTec. These results validate the effectiveness of introducing normal sample features as prior knowledge. Compared with Cutpaste, SGSF can be found to be more competitive. It shows the effectiveness of SAM to generate more effective forged anomalous samples as well as to provide guidance information.

In Table \ref{TAB.2}, we compare the anomaly classification performance of SGSF and unsupervised, weakly, and supervised anomaly detection methods on DAGM dataset.

The experimental results show that the proposed method outperforms the unsupervised and weakly supervised methods in 10 and 6 categories, indicating the superiority of the proposed method. The proposed method  achieves AUROC 99.9\%, indicating that SGSF can solve the anomaly classification challenge of grayscale images well. Specifically, the performance of SGSF classification without real anomaly supervision information is the same as the supervised methods. Importantly, there is a large distribution gap between the forged anomalous samples and the grayscale anomalies of DAGM, yet the anomalous samples in DAGM can be detected. It demonstrates that SGSF can address the challenge of distribution differences causing model to be inapplicable to real anomalies.

In Table \ref{TAB.3}, we compare the anomaly classification performance of SGSF and unsupervised and weakly anomaly detection methods on the KolektorSDD2 dataset. 
The experimental results show that SGSF achieves the best performance for the weakly supervised methods and unsupervised methods. SGSF outperforms unsupervised methods U-std 21\% and F-AnoGan 30\%, indicating the effectiveness of introducing normal features to guide segmentation network localization in unsupervised setting.

\emph{\textbf{(2) Anomaly localization: }}
To evaluate the performance of SGSF in anomaly localization task, we conduct experiment on the MVTec dataset. The experimental results of the proposed method and unsupervised anomaly detection methods are shown in Table \ref{tab4}. The proposed SGSF achieves mean $AUROC_{loc}$ 97.2\% and  mean $Ap_{loc}$ 73.8\%. DRAEM \cite{zavrtanik2021draem} points out that the $Ap_{loc}$ metric is able to evaluate the performance of the model in detecting normal and abnormal regions, and $AUROC_{loc}$ is influenced by the large number of normal pixel localization results. The proposed method achieved similar results with PaDim and DRAEM in $AUROC_{loc}$. However, SGSF is 18\% higher than PaDim and 5\% higher than DRAEM in $AP_{loc}$. The remaining unsupervised anomaly localization methods have similar phenomenon, while it is difficult to obtain good localization performance for anomalous regions. Due to the fact that the segmentation module of SGSF has the prior knowledge of normal patterns, it can detect all patterns different from normal as anomalies, which is not available in other methods. In sumary, the proposed SGSF method achieves the state-of-the-art localization results. 

In Figure \ref{fig:compare}, we visualize the anomaly localization results of the proposed method and the DRAEM.
 It can be seen that SGSF has the following advantages. First, SGSF is more accurate for detection of normal regions in the abnormal samples, as shown in (1), (5), (6), and (10). The localization map of DRAEM has many mislocalized anomalous regions. SGSF directly utilizes normal sample features as guidance features, thus ensuring the validity of the guidance information. Therefore, the segmentation module can correctly identify the normal features and segments the abnormal features that are different from the normal features. Second, SGSF has better performance for detail boundary localization, as shown in (3) and (8). It demonstrates the advantage of using multi-scale features for guidance in SGSF. The low-level features are effective for providing effective guidance about the boundary information, resulting in excellent localization performance of the segmentation module. Third, (9), (11), and (12) show that our method has advantages in localizing tiny anomalous regions. The detail feature information acts as guidance features that allow the model to effectively locate small anomalies. Fourth, (2), (7), and (15) show that DRAEM is wrong in predicting the background. SGSF addresses this problem in the Saliency Augmentation module. SAM introduces saliency map that make the model more focus on foreground content. However, Figure \ref{fig:figure3} (C) demonstrates that SGSF can also precisely localize to anomalies outside the region of saliency.
Finally, the SGSF is more sensitive for locating anomalous regions. As shown in (13), the anomalies are subtle textural differences. SGSF can still accurately locate detailed anomalies, which are not available with DRAEM.

Although SGSF achieves good results in most categories, it does not perform well in the ``transistor" category. Figure \ref{fig:tu7} shows the localization results of the samples in ``transistor" category. Groundtruth of some abnormal samples are marked not only with ``component bending'' but also with semantically missing regions. The model is required to further understand the fine-grained semantics of normal samples: the position of the component is related to whether the sample is abnormal or normal. Thus, the PaDim method of modeling images by patch location can better locate anomalies in the ``transitior'' category.  SGSF only locates the location of the anomalous component, but does not predict the original component location.

\begin{table}[]
	\centering
	
	\renewcommand\arraystretch{1.5}
	\caption{The impact of model size on the performance of the methods. the DRAEM$^1$,DRAEM$^2$ use the original structure and the network structure of this paper, respectively.}
	
	\setlength{\tabcolsep}{1.3mm}{
		\begin{tabular}{@{}c|cc|c|ccc@{}}
			\hline
			\multicolumn{1}{c|}{Method} & \multicolumn{2}{c|}{NET} & Size & \multicolumn{3}{c}{Metric} \\ \hline
			& Original     & SGSF     &   & AUROC$_{det}$   & AUROC$_{loc}$   & AP$_{loc}$     \\ \hline
			DRAEM$_1$  & \checkmark       &     &    97M            & 97.9    & 97.3    & 68.3   \\
			DRAEM$_2$  &              & \checkmark         & 65M  & 94.2    & 94.4    & 55.4   \\
			SGSF   &              & \checkmark         & 70M  & 96.6    & 97.2    & 73.8   \\ \hline
		\end{tabular}
	}\label{Table6}
\end{table}
\subsection{Ablation Study}
\begin{table}[]
	\centering
	\renewcommand\arraystretch{1.5}
	\caption{The impact of the number of samples $n$ in the contrast set on the SGSF.  ``W/O. A'' indicates whether the contrast set contains the input images.}
	
	\setlength{\tabcolsep}{2.5mm}{
		\begin{tabular}{c|cc|ccc}
			\hline
			&	\multicolumn{2}{c|}{Setting} & \multicolumn{3}{c}{Metric} \\ \hline
			&	$n$           & W/O. A.         & AUROC$_{det}$   & AUROC$_{loc}$   & AP$_{loc}$     \\ \hline
			
			(1)&$n=2$&-&96.9&96.5&72.5\\
			(2)&$n=4$         &  -           & 96.6    & 97.2    & 73.8   \\
			(3)&$n=5$         & \checkmark           & 97.0    & 97.0    & 72.2   \\
			(4)&$n=9$         & -           & 95.8    & 96.4    & 72.8  \\ 
			\hline
		\end{tabular}
		\label{Table7}
	}
\end{table}

\par 
\emph{\textbf{(1) Self-supervised Module.}}
To validate the effectiveness of the self-supervised module, ablation experiments were conducted without self-supervised module. In other words, the segmentation module directly segments forged anomalous and normal regions without guidance features as a priori knowledge. The results are shown in Table \ref{Tabel5} (1), and the result of proposed method is shown in Table \ref{Tabel5} (SGSF). It can be clearly found that the anomaly detection performance decreases with the lack of a self-supervised module. Particularly, the AP metrics decrease by 17\% compared to the method that introduces a self-monitoring module. Even though the SAM can generate forged anomalous samples close to the real anomalies to participate in training, there is still a distribution gap with the real anomalous samples. The self-supervision module is a key part of SGSF and the guidance features it provides are very important. The segmentation module is guided by the prior knowledge to segment anomalous regions only under the condition of understanding the normal pattern. Without normal feature information to guide, the performance is poor.
 The experimental results validate the effectiveness of the self-supervised module.
\par
\emph{\textbf{(2) Saliency Perlin Noise Generator. }} To validate the effectiveness of the proposed SPNG, experiments were conducted to explore the differences between it and traditional generation methods. Figure \ref{fig:figure3} (A) shows the forged anomalous samples generated by the conventional Perlin Noise Generator (PNG), while Figure \ref{fig:figure3} (B) shows the proposed Saliency Perlin Noise Generator (SPNG). Obviously, SGSF focuses on saliency regions to generate anomalies, while DRAEM tends to generate more noisy samples in the background, especially in the category of small foreground objects (e.g., ``screw"). Table \ref{Tabel5} (2) shows the performance using PNG. Table \ref{Tabel5} ``(SGSF)" shows the results of using SPNG. The results of both anomaly localization and anomaly classification are decrease, which validates the effectiveness of SPNG. SPNG reduces the noise sample generation and improves the learning efficiency of effective samples. Specifically, although SGSF produces anomalies only in the saliency region, however, we can still obtain good localization results on regions outside the saliency region, such as the zipper shown in Figure \ref{fig:figure3} (C) (the saliency map contains only the ``zipper" not the ``cloth", but anomalies on the ``cloth" can be detected).

\begin{table}[]
	\centering
	\centering
	\renewcommand\arraystretch{1.5}
	\caption{Ablation study with adaptive segmentation strategy. $b = 0$ indicates the fixed threshold is utilized.}
	
	\setlength{\tabcolsep}{5mm}{
		\begin{tabular}{@{}cc|ccc@{}}
			\hline
			\multicolumn{2}{c|}{Setting} & \multicolumn{3}{c}{Metric} \\ \hline
			a            & b            & AUROC$_{det}$   & AUROC$_{loc}$   & AP$_{loc}$   \\\hline
			0.2          & 0            & 95.7   & 96.4   & 68.3  \\
			0.2          & 0.2          & 95.2   & 95.7   & 71.6  \\
			0.4          & 0            & 94.3   & 96.1    & 69.7   \\
			0.4          & 0.2          & 96.6    & 97.2   & 73.8   \\
			0.6          & 0            & 95.8   & 94.8   & 71.5  \\ \hline
	\end{tabular}}
	\label{Table8}
\end{table}

\begin{table}[]
	\centering
	\renewcommand\arraystretch{1.5}
	\caption{Ablation study of importance parameter $\lambda$ in loss functions.}
	
	\setlength{\tabcolsep}{5mm}{
		\begin{tabular}{@{}c|c|ccc@{}}
			\hline
			& Setting & \multicolumn{3}{c}{Metric} \\ \hline
			& $\lambda$     &AUROC$_{det}$   & AUROC$_{loc}$   & AP$_{loc}$   \\\hline
			(1) & 0.01    & 95.4    & 94.8    & 67.4   \\
			(2) & 0.1     & 95.9    & 94.9    & 66.4   \\
			(3) & 1       & 96.6    & 97.7    & 73.8   \\
			(4) & 2       & 95.8    & 95.1    & 70.0   \\ \hline
	\end{tabular}}
	\label{Table9}
\end{table}
\par 
\emph{\textbf{(3) Reconstruction task and denoising task. }} 
To validate the influence of self-supervised tasks in the self-supervised module, different self-supervised tasks are selected to validate the performance.
Theoretically, common self-supervised tasks are applicable to the proposed framework. In this paper, we experimentally explore the reconstruction task and the denoising task. Specifically, we replace the self-supervised task of the self-supervised module with either the reconstruction task or the denoising task to validate their effects on the performance of anomaly detection. The input of reconstruction model is randomly sampled images in contrast set, and the input of denoising model is forged anomaly samples, the network structures are the same and both are updated by the loss function of the self-supervised module.

The experimental results are shown in Table \ref{Tabel5} for ``SGSF" and (3). The results show that the anomaly classification performance decreases by 2\% using denoising task compared to reconstruction task, and the anomaly localization in $AUROC_{loc}$ and $AP_{loc}$ decreases by 3\% and 7\%, respectively. Therefore, the reconstruction task is more suitable for SGSF. The reason is that SGSF relies on the guidance features as the prior knowledge to guide the segmentation module to locate anomalous regions. Therefore, it is important to ensure that the guidance features are noise-free normal features. The input of reconstruction network are normal samples, while the input of the denoising network are forged anomalous samples. Obviously, the features extracted from the reconstruction network are noise-free and are more suitable for the guidance features. Although the denoising network can denoise some anomalous regions by training, it is difficult for a simple denoising network to guarantee a good denoising effect for all anomalous regions. Therefore, the guidance features extracted from the denoising network may be noisy anomalous features and cannot serve as guidance. In summary, the choice of reconstruction task can achieve excellent anomaly detection performance.

\par\emph{\textbf{(4) Comparison of the proposed method with DRAEM. }} To demonstrate the performance and parameter advantages of our method and DRAEM, we design three different settings. The setting of experiment DRAEM$_1$ uses the original network structure. The setting of experiment DRAEM$_2$ uses the network structure of SGSF. The difference between the original network and SGSF's network are that our network is shallower and has smaller parameters. The experimental results are shown in \ref{Table6}. It show that our method has better localization results than DRAEM$_1$ and the parameters is reduced by $1/3$, which shows that SGSF is more efficient and accurate. Second, after changing the network of DRAEM to the network structure of this paper, the anomaly classification and localization results are reduce by 3.7\%, 2.3\% and 12.9\%, respectively. By comparing DRAEM$_2$ with the proposed method, it is shown that our method is more competitive. The smaller model brings more competitive localization and classification results.

\par\emph{\textbf{(5) Contrast set. }}
 To validate the effect of the number of images $n$ in contrast set, ablation study with different  values of $n$ were explored. The results are shown in Table \ref{Table7}. ``W/O.A." indicates whether contrast set contains the original image $A$. Table \ref{Table7} (2) and (3) validate the effect of containing itself $A$ in the contrast set on anomaly detection. It can be observed that the performance of anomaly detection containing $A$ is lower than the performance without $A$. The inclusion of the original image in the contrast set makes the segmentation module in SGSF overly focused on the guidance of low-level features and produces a negative effect. From the results of Table \ref{Table7} (1), (2) and (4), the localization performance with $n=4$ exceeds remaining settings. Therefore, $n$ in the contrast set is set to $4$ without the original image. Large $n$ value leads to a larger difference between the image in the contrast set and the input image, which makes to guide information may be noisy. The noisy guidance features make the prior knowledge is invalid and the segmentation module localization effect will be reduced. When $n$ is too small, the segmentation module accepts less variation in the guidance features, which weakens the generalization ability of the segmentation module.

\par\emph{\textbf{(6) Adaptive threshold segmentation strategy. }}
To validate the effectiveness of the proposed adaptive threshold segmentation strategy, ablation experiments on $a$ and $b$ in Equ. (\ref{eq:1}) are conducted. The results are shown in Table \ref{Table8}. The experimental results show that the anomaly detection results obtained with a fixed threshold are lower than the the adaptive threshold scheme. When $a=0.2$ and $b=0.2$, the localization result is obviously higher than those of the $a=0.2,b=0$ setting. The same conclusion is obtained for $a=0.4$. When the value of $a$ becomes larger, the obtained mask $M$ region becomes very numerous and small anomalous regions. After the encoder downsampling will make the anomalous features not obvious, which are not beneficial for model learning. The threshold value is fixed and the samples formed have little variation. Although good results can be achieved, the adaptive threshold scheme is more likely to realize the potential of SGSF. The results show that when the threshold is set above 0.6, the segmented region becomes smaller and scattered and the performance decreases. The best results in the table are $a=0.4$ and $b=0.2$.

\par\emph{\textbf{(7) Ablation study of the $\lambda$. }}
To validate the effect of hyperparameters $\lambda$ in the loss function on the model, four sets of comparison experiments with hyperparameters $\lambda$ were conducted. In the experiment $\lambda$ is set to 0.01, 0.1, 1, and 2, respectively.
 The experimental results are shown in Table \ref{Table9}. When $\lambda=0.01$, it indicates that the loss of the self-supervised module is smaller than the overall loss. From the results in Table \ref{Table9}, it is found that the performance of both localization and classification decreases, which may be due to the loss function causing the supervised module to not fully converge. The self-supervised module plays a very important role in this paper. The reconstruction effect of the self-supervised module is not significant, then there may be noise in the extracted guidance features, which affects the learning effect of the segmentation module. Therefore, the value of $\lambda$ cannot be too small. When $\lambda$ value is large, the performance of the model starts to degrade. Therefore, the optimal hyperparameter $\lambda = 1$ is chosen in this paper.

\section{Conclusion}
In this paper, a novel unsupervised method is proposed for visual anomaly detection by exploring the Saliency Augmentation Module, Self-supervised Guided Network, and the segmentation module. SAM generates saliency Perlin noise utilizing saliency maps generates forged anomaly samples that are more suitable for model learning. SGN uses the guidance features extracted in the self-supervision module, then the guidance fusion module fuses the guidance features with the features extracted by the segmentation module, finally, the segmentation module along with the guidance features guides to segment the anomalous features that are different from the guidance features. Competitive results are achieved on several benchmark datasets, especially in anomaly localization. The ablation studies validate the effectiveness of each module. In the future, we will gradually address the challenge of low accuracy of existing models in ``transistor" category localization and propose more reasonable alignment methods to make the guidance features more useful.

{
\bibliographystyle{IEEEtran}
\bibliography{reference}
}

\begin{IEEEbiography}[{\includegraphics[width=1in,height=1.25in,clip,keepaspectratio]{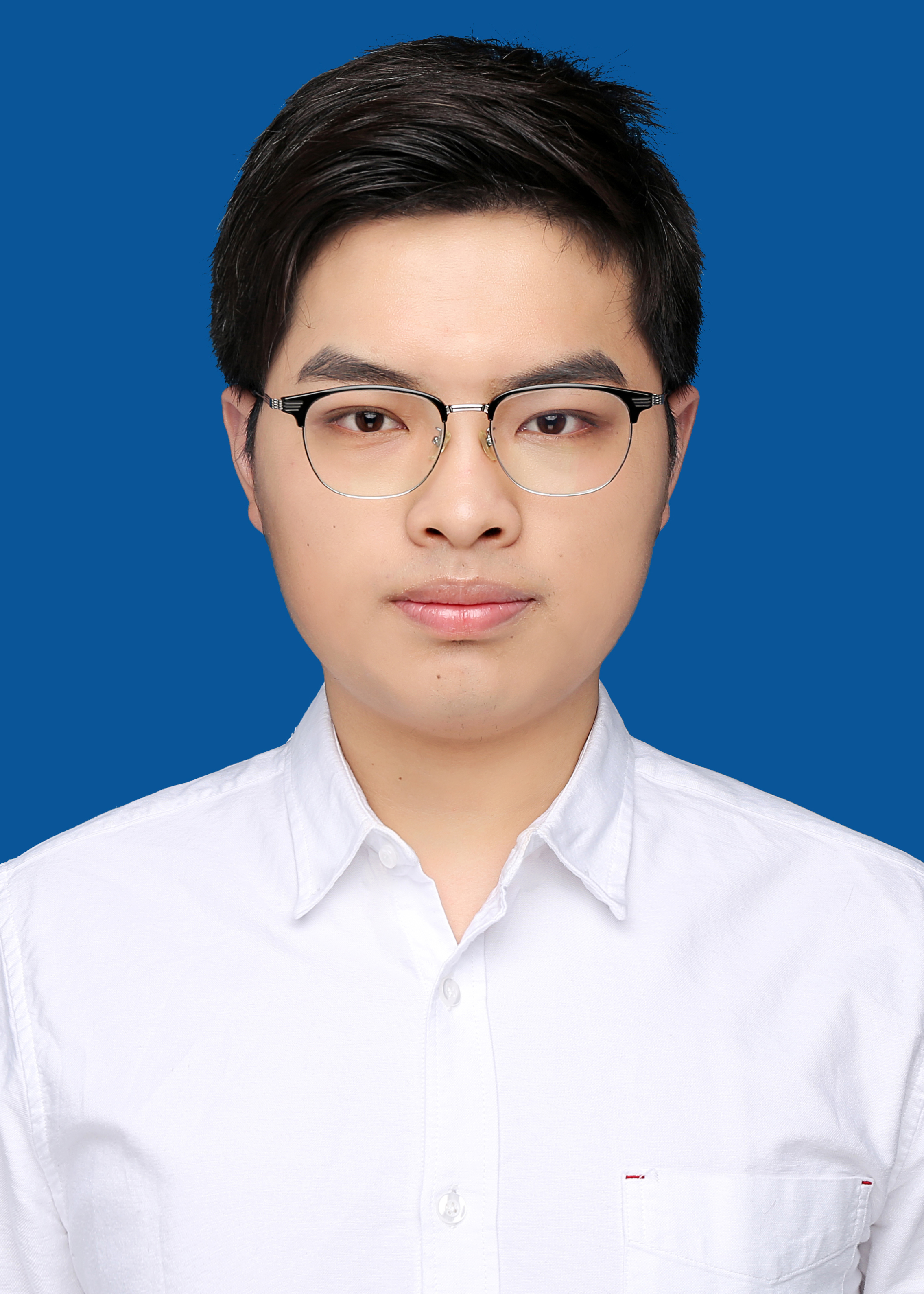}}]{Peng Xing} is currently pursuing the master’s
	degree with the School of Computer Science and
	Engineering, Nanjing University of Science and
	Technology. His current
research interests include anomaly detection and
unsupervised deep learning.
\end{IEEEbiography}

\vspace{11pt}
\begin{IEEEbiography}[{\includegraphics[width=1in,height=1.25in,clip,keepaspectratio]{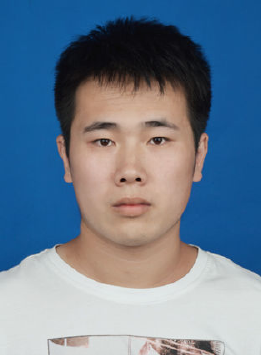}}]{Yanpeng Sun} received the MS degree at Guilin University Of Electronic Technology, China, in 2019. He is currently pursuing the Ph.D. degree with the School of Computer Science and Engineering, Nanjing University of Science and Technology, China. His research interests include deep learning, visual segmentation and understanding, etc.
	
\end{IEEEbiography}
\vspace{11pt}
\begin{IEEEbiography}[{\includegraphics[width=1in,height=1.25in,clip,keepaspectratio]{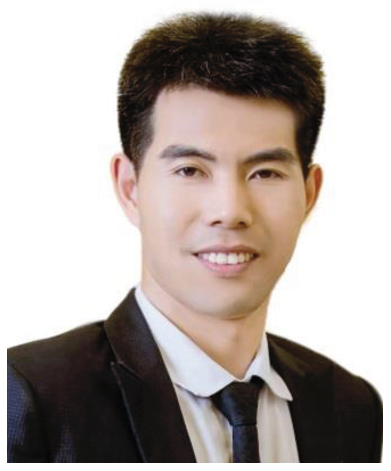}}]{Zechao Li} is currently a Professor at the Nanjing University of Science and Technology. He received his Ph.D degree from National Laboratory of Pattern Recognition, Institute of Automation, Chinese Academy of Sciences in 2013, and his B.E. degree from the University of Science and Technology of China in 2008. His research interests include big media analysis, computer vision, etc. He was a recipient of the best paper award in ACM Multimedia Asia 2020, and the best student paper award in ICIMCS 2018. He serves as an Associate Editor for IEEE \scshape Transactions on Neural Networks and Learning Systems.
\end{IEEEbiography}

\vfill

\end{document}